\documentclass{article}
\usepackage{spconf,amsmath,graphicx}

\usepackage{times}

\usepackage{epsfig}
\usepackage{booktabs}
\usepackage{algorithmic}
\usepackage{textcomp}

\usepackage{framed,multirow}
\usepackage{xcolor}
\usepackage{amssymb}

\title{RL-LOGO: Deep Reinforcement Learning Localization \\for Logo Recognition}
\name{
Masato Fujitake
}
\address{
FA Research, 
Fast Accounting Co., Ltd.
Japan \\
fujitake@fastaccounting.co.jp
}

\begin{document}
\maketitle
\begin{abstract}
This paper proposes a novel logo image recognition approach incorporating a localization technique based on reinforcement learning.
Logo recognition is an image classification task identifying a brand in an image.
As the size and position of a logo vary widely from image to image, it is necessary to determine its position for accurate recognition.
However, because there is no annotation for the position coordinates, it is impossible to train and infer the location of the logo in the image.
Therefore, we propose a deep reinforcement learning localization method for logo recognition (RL-LOGO).
It utilizes deep reinforcement learning to identify a logo region in images without annotations of the positions, thereby improving classification accuracy.
We demonstrated a significant improvement in accuracy compared with existing methods in several published benchmarks. 
Specifically, we achieved an 18-point accuracy improvement over competitive methods on the complex dataset Logo-2K+.
This demonstrates that the proposed method is a promising approach to logo recognition in real-world applications.

\end{abstract}
\section{Introduction}
\label{sec:intro}
Logo recognition identifies a brand associated with a logo in a natural scene and web image. 
It has become increasingly important because it provides valuable insights for various applications, including the challenges of advertising analysis~\cite{wang2020logo2k}, document~\cite{fujitake2023a3s} and scene~\cite{fujitake2021tcbam}.
Since logo recognition aims to estimate the brand category corresponding to an image, previous studies have treated it as a standard image classification task using convolutional neural networks (CNNs).

\begin{figure}[htp]
   \centering
   	\includegraphics[width=0.70\columnwidth, keepaspectratio]{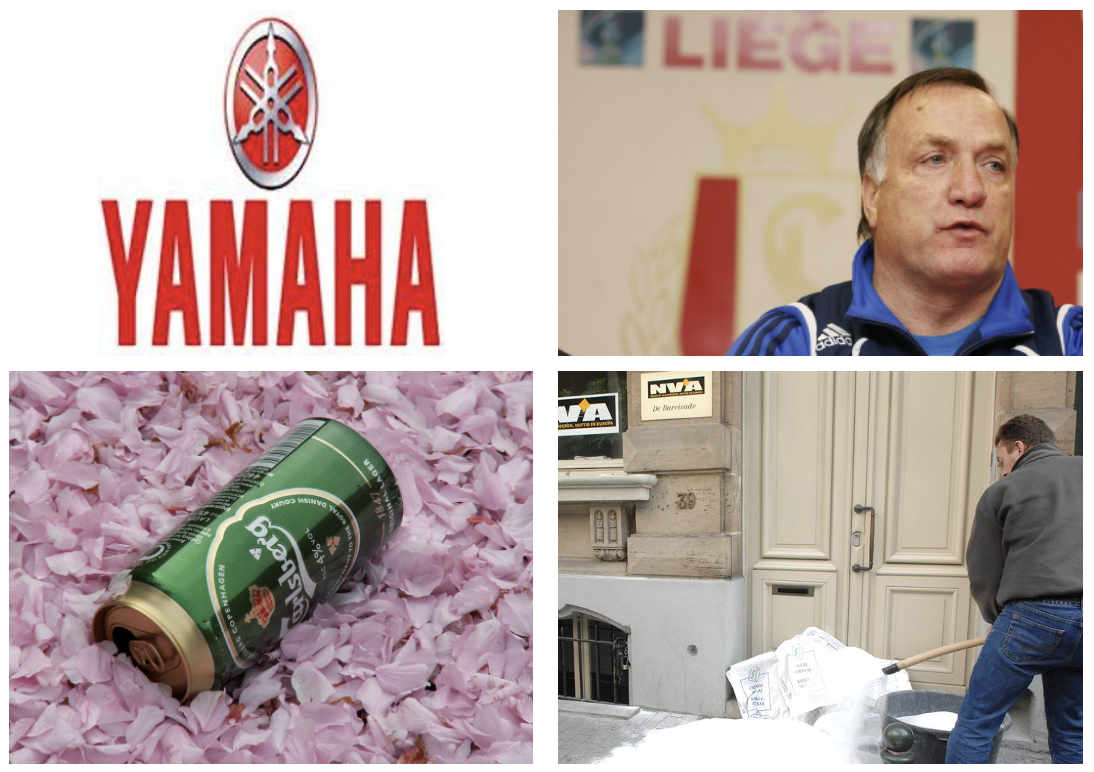}
   \caption{
Example of logo recognition from the benchmark datasets.
Logos are included in the image, but unlike typical image recognition, their position and size differ significantly depending on the image, making recognition difficult.
}
   \label{fig:sample_logocls}
       \vspace*{-1.00\baselineskip}
\end{figure}

However, recognizing logos is a specialized task that requires an approach different from general image classification. 
This is because logo images have unique characteristics that make applying a simple image classification method challenging. 
In general image classification tasks~\cite{deng2009imagenet}, the object to be classified, such as a dog, is typically presented at the center of the image.
In contrast, as shown in Figure~\ref{fig:sample_logocls}, the position and size of logos in logo recognition images vary significantly from image to image.
As a result, existing methods~\cite{yang2018ntfnet, wang2020logo2k} that use the entire image for classification often suffer from problems such as background noise and scale variation, leading to undesirable performance.

We propose a novel logo recognition method, RL-LOGO, based on deep reinforcement learning.
To accurately recognize a logo in an image, it is crucial to narrow down its location in the image. 
Our approach employs deep reinforcement learning to iteratively refine, recognize, and determine potential regions where the logo might be located in the image.
As a general object localization task, localization using deep reinforcement learning in images has been studied~\cite{caicedo2015drl_active}.
However, the method requires object location annotation to compute the reward.
Since the logo's position annotations are not available for training and inference, we propose a confidence-guided reward function that uses the confidence score of image classification for localization.
Our approach effectively addresses issues such as scale variation by performing image classification on candidate logo regions, resulting in highly accurate logo recognition. 
To validate our method, we conducted comprehensive experiments and compared it with state-of-the-art logo recognition techniques on multiple benchmark datasets.
The results show that our proposed method outperforms existing methods, highlighting its potential for real-world applications.
The main contributions of this study are as follows:
\begin{itemize}
    \item We present a novel logo recognition method using deep reinforcement learning, RL-LOGO, which distinguishes itself from conventional logo recognition techniques.
    \item RL-LOGO has achieved state-of-the-art benchmark results despite its simple but effective reward function.
\end{itemize}

\section{Related Works} \noindent
\textbf{Logo Recognition:} 
Logo recognition is an active research area in computer vision, and numerous techniques have been proposed owing to its importance in real-world applications. 
Traditional approaches have employed handcrafted feature descriptors, whereas the emergence of deep learning has led to CNNs becoming the standard for logo recognition. 
Several studies have adapted and fine-tuned pre-trained CNN models for logo recognition tasks~\cite{he2019bagimgcls}, such as VGG~\cite{simonyan2014vgg} and ResNet~\cite{he2016resnet}. 
Others have utilized attention mechanisms~\cite{yang2018ntfnet} and data augmentation~\cite{wang2020logo2k} to enhance performance. 
However, these methods still struggle with noisy backgrounds and scale variation. 
Our method distinguishes itself by addressing these challenges by dynamically determining the regions the agent should focus on without explicit localization annotation.

\noindent
\textbf{Logo Detection:} 
Logo detection has gained popularity in recent years~\cite{sujuan2023logodetection, wang2022logodet, romberg2011flickrlogos}. 
It involves identifying the position and category of a logo within an image.
Various methods have been proposed to improve accuracy~\cite{wang2022logodet, fehervari2019scalable} including text recognition~\cite{fujitake2023diffusionstr, fujitake2024dtrocr}, but annotations for logo detection are expensive, making it challenging for real-world practical applications. 
Thus, this study focuses on logo recognition and analyzes our method using a logo detection benchmark dataset~\cite{romberg2011flickrlogos}.

\noindent
\textbf{Deep Reinforcement Learning for Object Localization:} 
Recently, some studies have applied deep reinforcement learning (DRL)~\cite{he2019gaze}, which merges deep learning and reinforcement learning to an object localization task, using Q-learning algorithms~\cite{mnih2015dqn} to direct the search process~\cite{caicedo2015drl_active}. 
However, such an approach requires the object's positional annotation in the training phase to provide feedback rewards. 
Our study is the first work to utilize DRL for logo recognition and proposes a new reward function to address the issues.

\label{sec:relatedwork}

\section{Method} \label{sec:method}
\begin{figure}[htp]
   \centering
   	\includegraphics[width=0.90\columnwidth, keepaspectratio]{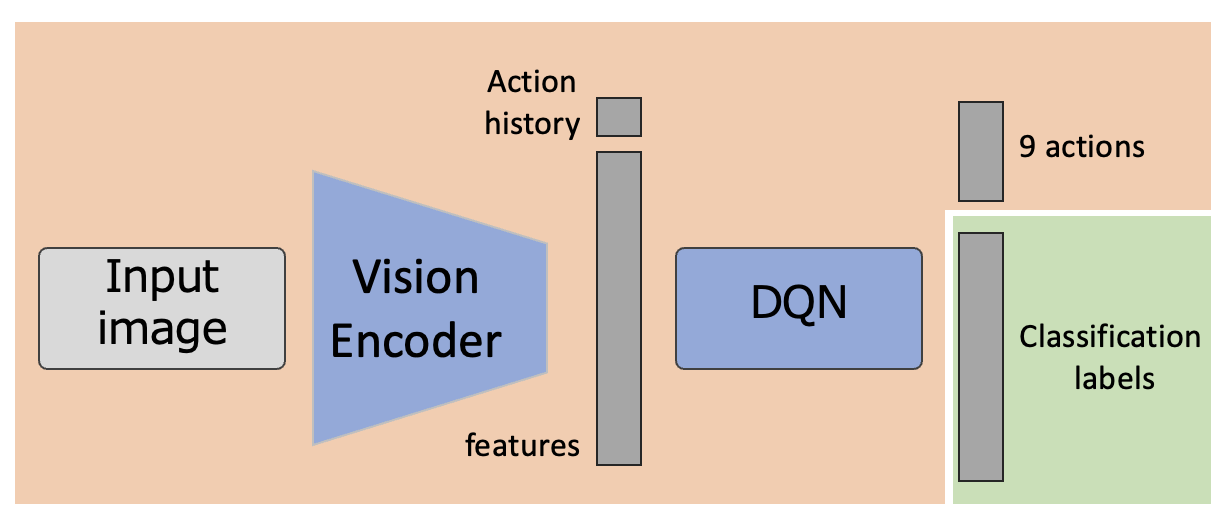}
   \caption{
   Architecture of the proposed RL-LOGO, which consists of a vision encoder and DQN module, performs position prediction and class recognition.
The orange area represents modifications based on previous research for object localization, and the green area represents our proposed method.
   }
   \label{fig:architecture}
       \vspace*{-1.00\baselineskip}
\end{figure}

The structure of the proposed method is shown in Figure~\ref{fig:architecture}.
It is simple, consisting mainly of a vision encoder that acquires features from images and a deep Q-network (DQN)~\cite{mnih2015dqn} that determines the following action and recognizes the logo.
Our proposed method extends the object localization method~\cite{caicedo2015drl_active} to the logo recognition task in two ways.
The first is an extension to the classification task.
The second is introducing a new reward function explicitly designed to localize the logo without location information.
We briefly review the prior work and describe our proposed method in detail.

\subsection{Object Localization using DQN}
A study has been proposed to predict the location of objects in an image using DQN~\cite{caicedo2015drl_active}.
It is a deep reinforcement learning algorithm that uses deep neural networks to approximate the relationship between states and actions.
The prior work~\cite{caicedo2015drl_active} has used elementary transformation actions to facilitate bounding box transformations, allowing agents to identify the most accurate location of target objects based on a top-down inference process.
More specifically, a visual encoder, such as CNNs, is employed to extract features from the input image, as shown in the orange area of Figure~\ref{fig:architecture}.
Next, the image features are flattened into a 1D sequence, and a previously selected action history is combined with them and used as input to the DQN.
It consists of two linear layers with 1,024 units to estimate nine actions.
These actions include moving the bounding box vertically and horizontally, scaling it up/down, changing its height and width aspect ratio, and determining its endpoint.
This process is repeated until the prediction of an endpoint.
As a reward function of the object's position estimation, the intersection point (IoU) over the sum of the object's ground-truth region and predicted region is calculated at each step.
The difference between the IoU of the current and previous steps is computed, and a positive reward is given if the object position is captured more accurately.
Subsequently, when the final position is determined, a final reward is given if the IoU of the predicted and ground-truth boxes exceeds a threshold value.
Mean squared error loss is used for the optimization.

\subsection{Extension to Logo Recognition}
Two issues must be solved to apply prior research to logo recognition.
The first is the prediction of the logo-class labels.
Second, it is necessary to establish a reward function for datasets lacking logo-position information.
To address the first issue, we adapted the agent's output to generate a 1-hot vector representing class labels along with the action outputs, as depicted in the green region of Figure~\ref{fig:architecture}. 
The number of classes depends on the dataset.
The categorical optimization is performed using cross-entropy loss used in image classification.
We propose introducing a simple but effective confidence-guided reward function for the second issue.

\subsubsection{Confidence-guided Reward Function}
The confidence-guided reward function serves as a reward metric that leverages the agent's confidence score of the target class label. 
Our hypothesis posits that, for an agent to achieve high confidence in logo recognition, it must be capable of obtaining logo information from optimal positions.
Conversely, if the agent focuses on off-target regions, the confidence associated with the target class should be low.
Building upon this notion, we have introduced a reward metric that quantifies the difference in the target class's confidence at each step. 
The reward function $R_a(s,s')$, when it chooses the action $a$ to move from state $s$ to $s'$, can be expressed as follows:

\begin{equation} \label{eq:reward_action}
    R_a(s,s') = sign\left( c' - c \right),
\end{equation}
where $c'$ and $c$ are the image's target class confidence, normalized by the softmax function over all classes.

Another reward function is also introduced to determine the validity of the positioning decision once the final position is determined.
The reward at the endpoint is a threshold function of the confidence score of the target class, as follows:

\begin{equation} \label{eq:reward_endpoint}
R_\omega(s,s') =
\left\{
	\begin{array}{ll}
		+\eta  & \mbox{if } c' \geq \tau \\
		-\eta & \mbox{otherwise },
	\end{array}
\right.
\end{equation}
where $\omega$ is the endpoint action, $\eta$ is the endpoint reward, and $\tau$ is the minimum confidence threshold for a positive detection.
We set $\eta$ and $\tau$ to $2.0$, $0.75$, respectively in our experiments.

\subsubsection{Training Procedure}
The learning process comprises two main stages: pre-training and joint training. 
During the pre-training phase, the agent is trained on a dataset comprising logo images, akin to a conventional image classification task. 
This stage facilitates estimating the confidence score, which is crucial for following joint training.

Subsequently, the joint training phase incorporates DRL to enable the simultaneous execution of localization and classification tasks as an agent~\cite{caicedo2015drl_active}.
The losses for action decision and logo class prediction are computed by dividing the agent's output.
All loss weights are equal.
This integrated approach allows for an efficient and comprehensive learning process.

\subsubsection{Inference Procedure}
During inference, the agent starts with the entire image and estimates and performs each action to identify the logo area.
When an endpoint decision is made, the category prediction results are used for recognition.
The maximum number of actions is set to 40 steps by previous research, in which case the last category prediction result is used.

\section{Experiments} \label{sec:experiments}

\begin{table*}[t]
  \caption{
  Logo recognition performance comparison on Loto-2K+, BelgaLogos, FlickrLogos-32, and WebLogo-2M.
  }
\centering
  \label{tab:method_overall_result}
\resizebox{1.40\columnwidth}{!}{%
\begin{tabular}{c|cc|cc|cc|cc}
\toprule
\multirow{2}{*}{Methods} & \multicolumn{2}{c|}{Logo-2K+} & \multicolumn{2}{c|}{BelgaLogos} & \multicolumn{2}{c|}{FlickrLogos-32} & \multicolumn{2}{c}{WebLogo-2M} \\
                         & Top-1         & Top-5          & Top-1       & Top-5        & Top-1         & Top-5           & Top-1 & Top-5            \\\midrule 
VGGNet-16~\cite{simonyan2014vgg}                   & 62.83      & 89.01 & $-$ & $-$ & $-$ & $-$ & 62.88 & 83.23\\
ResNet-50~\cite{he2016resnet}                      & 66.34      & 91.01 & $-$ & $-$ & $-$ & $-$ & 62.93 & 83.32\\
ResNet-152~\cite{he2016resnet}                     & 67.65      & 91.52 & $-$ & $-$ & $-$ & $-$ & $-$ & $-$\\
Efficient(ResNet-50)\cite{he2019bagimgcls}         & 66.94      & 91.30 & $-$ & $-$ & $-$ & $-$ & $-$ & $-$\\
Efficient(ResNet-152)~~\cite{he2019bagimgcls}        & 67.99      & 91.68 & $-$ & $-$ & $-$ & $-$ & $-$ & $-$\\
NTS-Net(ResNet-50)~\cite{yang2018ntfnet} & 69.41      & 91.95 & 93.33 & 96.15 & 94.14 & 96.29 & 63.67 & 84.31\\
DRNA(ResNet-50)~\cite{wang2020logo2k}    & 71.12      & 92.33 & 94.44 & 97.11 & 95.33 & 97.17 & 64.82 & 86.12\\
DRNA(ResNet-152)~\cite{wang2020logo2k}   & 72.09      & 93.45 & 95.82 & 98.40 & 96.63 & 98.80 & $-$ & $-$\\

\midrule
Ours(ResNet-50) & 89.43  & 95.32  & 97.49 & 99.03 & 98.10 & 99.16 & 77.38 & 96.85\\
Ours(ResNet-152) & \textbf{90.29}  & \textbf{96.15}  & \textbf{98.98} & \textbf{99.71} & \textbf{98.94} & \textbf{99.67} & \textbf{78.25} & \textbf{97.31}\\
 \bottomrule
\end{tabular}
}
    \vspace*{-1.00\baselineskip}
\end{table*}

\begin{figure}[htp]
   \centering
   	\includegraphics[width=0.75\columnwidth, keepaspectratio]{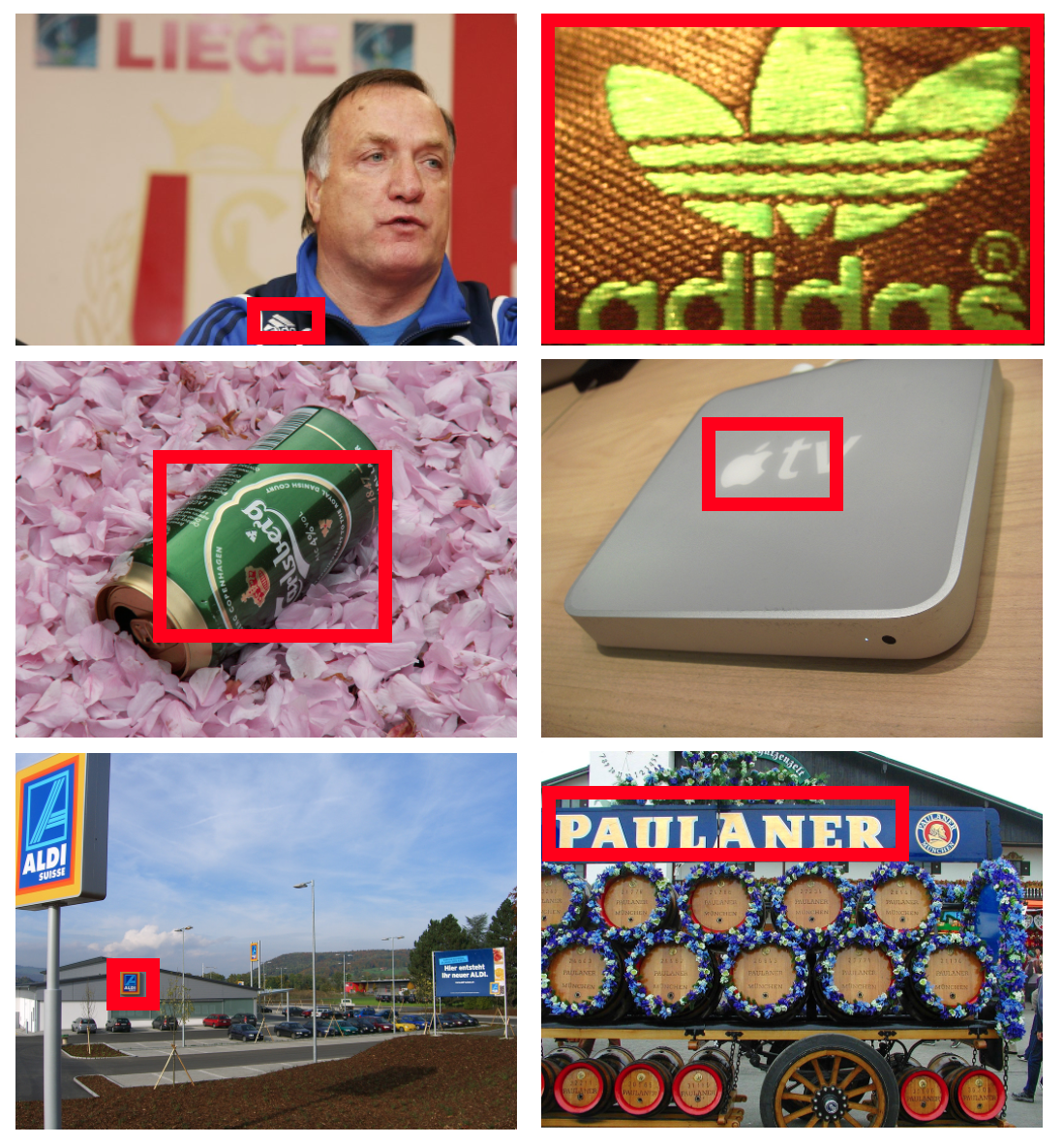}
   \caption{Visualization of logo localization.
The proposed method finally recognized the logo at the position indicated by the red frame in the image.
The proposed method locates and recognizes logo brands.
   }
   \label{fig:visualization_result}
       \vspace*{-1.00\baselineskip}
\end{figure}

\begin{table}[tb]
\centering
\caption{
Action iteration statistics. 
}
\label{tab:iteration_analysis}
\small
\begin{tabular}{c|c|c }
\toprule
      & Median  & Mean \\ \hline
Iteration & 0.4  & 4.9  \\
\bottomrule
\end{tabular}
       \vspace*{-0.50\baselineskip}
\end{table}

\begin{table}[tb]
\centering
\caption{
Impact of reward function to localization.
}
\label{tab:reward_function_analysis}
\resizebox{0.8\columnwidth}{!}{%
\small
\begin{tabular}{c|c|c }
\toprule
Method      & Positional annotation  & Recall [\%] \\ \hline
IoU-based~\cite{caicedo2015drl_active} & \checkmark  & 62.3  \\
Confidence-guided (ours) &   & 60.1  \\
\bottomrule
\end{tabular}
}
       \vspace*{-1.00\baselineskip}
\end{table}

\subsection{Dataset and Evaluation}
We evaluated our proposed method using public benchmarks. 
In line with previous research~\cite{wang2020logo2k}, we randomly split all datasets into train and evaluation by 70\% and 30\%, respectively.
The evaluation metric is Top-1 and Top-5 accuracies in image classification~\cite{he2016resnet}.
The datasets used are as follows:

\noindent
\textbf{Logo-2K+~\cite{wang2020logo2k}:} 
consists of 2,341 logo classes, a sizeable other class, and 167,140 images compared to other datasets.
It has variations, such as logo images in scenes and on the Web.

\noindent
\textbf{BelgaLogos~\cite{alexis09belgalogos}:} contains 37 product brand logo classes and 10,000 total images.
It is a small-sized logo data set.

\noindent
\textbf{FlickrLogos-32~\cite{romberg2011flickrlogos}:} includes 32 logo brands and 8,240 logo images from Flickr posted images.

\noindent
\textbf{WebLogo-2M~\cite{su2017weblogo}:} comprises 1,867,177 logo images across 194 different logo classes. 
It is considerably noisy due to unsupervised annotation.

\subsection{Implementation Details}
We used ResNet-50 and ResNet-152~\cite{he2016resnet}, pre-trained in ImageNet, as the vision encoder to make a fair comparison with previous studies.
The input images are resized to 224 pixels in height and width.
In the pre-training step, the model was optimized as typical image classification, using stochastic gradient descent with a momentum of 0.9, batch size of 256, and weight decay of 0.0001.
The model is trained for 100 epochs with an initial learning rate of 0.001 and reduced to 0.0001 after 20 epochs.
We initialize and train the model with zeros concerning action history and action.
As for data augmentation, random image rotation processes of 0, 90, 180, and 270 degrees were added to typical image classification ones~\cite{he2016resnet} because logos are sometimes rotated.

In the joint training, the agent is initialized with the weights learned in the pre-training, and a 15-epoch localization is learned using a greedy training strategy following previous research~\cite{caicedo2015drl_active}.
We also followed previous studies for bounding box operation setup.
In the first five epochs, the agent is linearly annealed from 1.0 to 0.1, and after the fifth epoch, it is fixed at 0.1.
All experiments are performed on four Nvidia four A100 GPUs in mixed precision using PyTorch.
We report an average score of a 4-shot experiment.

\subsection{Comparison with state-of-the-arts}
Table~\ref{tab:method_overall_result} presents the results of the proposed method and previous studies for each dataset. 
Our method surpasses the state-of-the-art performance on all datasets. 
It outperforms the strongest method, DRNA~\cite{wang2020logo2k}, which employs an attention mechanism for powerful feature regions.
In particular, we demonstrated the effectiveness of our approach on the Logo-2K+ and WebLogo-2M datasets with Top-1 accuracy improvements of 18.31 and 12.56 points, respectively, over DRNA.
Our method effectively directs the logo localization process, enabling the agent to focus on relevant regions for classification. 
This makes the method robust against background noise and scale variations in logo recognition.

\subsection{Detailed Analysis}
We performed a detailed analysis with the ResNet-50 agent.

\noindent
\textbf{Logo Localization Visualization:} 
Figure~\ref{fig:visualization_result} shows the bounding box with a red frame when logo recognition was determined.
The proposed method can generally identify the positions of logo to recognize them accurately.
However, some problematic cases exist, such as when multiple logos exist in an image, or when the logo symbol and text exist separately, which are difficult to judge.

\noindent
\textbf{The Number of Actions Until a Recognition:} 
Table~\ref{tab:iteration_analysis} is the statistics of how many times the proposed method performed actions before outputting the result of logo recognition on the Logo-2K+ validation set.
The number of iterations is zero if the logo is classified in the entire initial image.
The difference between the median and mean confirms the step depends on an image to recognize a logo.
The median is low because many logo images are reflected in the whole image, and the mean is 4.9 because the number of iterations increases slightly for small logos in an image.

\noindent
\textbf{Effect of Reward Functions:} 
We checked the accuracy of logo localization for different reward functions.
Since the previous study requires positional annotations when determining the reward value, we used FlickerLogos-32 annotations for logo detection and evaluated the localization performance, ignoring the class.
Table~\ref{tab:reward_function_analysis} displays the methods and the corresponding recall accuracies.
The proposed method with no positional annotation achieved comparable accuracy to the previous study, which required them.
This confirms the effectiveness of our method.

\section{Conclusion}\label{sec:conclusion}
This paper presented a novel approach that combines localization techniques to handle logo recognition with high variability in size and position.
We proposed a confidence-guided reward function with deep reinforcement learning to predict logo candidate regions. 
It can identify logo regions without requiring positional annotations, making it a valuable tool for logo recognition. 
Experiments on benchmark datasets have shown that our technique is superior to existing state-of-the-art methods and has great potential for practical applications.

\clearpage

\small{
\bibliographystyle{IEEEbib}
\bibliography{article}
}

\end{document}